\documentclass{article}

\usepackage{microtype}
\usepackage{graphicx}
\usepackage{subfigure}
\usepackage{booktabs} 

\usepackage{hyperref}

\usepackage[accepted]{icml2024}

\usepackage{amsmath}
\usepackage{amssymb}
\usepackage{mathtools}
\usepackage{amsthm}

\usepackage[capitalize,noabbrev]{cleveref}

\theoremstyle{plain}

\theoremstyle{definition}

\theoremstyle{remark}

\usepackage[textsize=tiny]{todonotes}
\usepackage{wrapfig}
\usepackage{nicefrac}
\usepackage{lipsum} 

\usepackage{listings} 
\usepackage{courier}
\definecolor{mygreen}{rgb}{0,0.6,0}
\definecolor{mygray}{rgb}{0.5,0.5,0.5}
\definecolor{mymauve}{rgb}{0.58,0,0.82}
\definecolor{superlightgray}{RGB}{240,240,240}
\definecolor{derekTableBlue}{RGB}{189,235,252}

\usepackage[labelfont=bf,format=plain]{caption}
\usepackage{subcaption}
\definecolor{darkBlue}{RGB}{0, 94, 184}
\definecolor{derekBlue}{RGB}{144,210,236}
\definecolor{forest}{RGB}{50,140,90}
\definecolor{derekTableBlue}{RGB}{189,235,252}
\definecolor{iglGreen}{RGB}{153,203,67}
\definecolor{coralRed}{RGB}{250,114,104}
\definecolor{gray}{RGB}{180,180,180}
\definecolor{orange}{RGB}{255,165,0}
\definecolor{TechnionBlue}{RGB}{8,33,78}
\definecolor{Purple}{RGB}{137, 99, 198}
\definecolor{lightgray}{gray}{0.65}

\newcommand*{\refsec}[1]{%
  \begingroup
    \def\sectionautorefname{Sec.}%
    \def\subsectionautorefname{Sec.}%
    \def\subsubsectionautorefname{Sec.}%
    \autoref{sec:#1}%
  \endgroup
}
\newcommand*{\refequ}[1]{%
  \begingroup
    \def\equationautorefname{Eq.}
    \autoref{equ:#1}%
  \endgroup
}

\newcommand*{\reffig}[1]{%
  \begingroup
    \def\figureautorefname{Fig.}%
    \autoref{fig:#1}%
  \endgroup
}

\DeclareMathOperator{\Tr}{Tr}

\newcommand{\R}{\mathbb{R}}
\newcommand{\C}{\mathbb{C}}
\newcommand{\M}{\mathcal{M}}
\newcommand{\Lc}{\mL_c}
\renewcommand{\i}{\imath}

\newcommand{\vecFont}[1]{\mathbf{#1}}

\def\ve{{\vecFont{e}}}

\def\vr{{\vecFont{r}}}

\def\vu{{\vecFont{u}}}
\def\vv{{\vecFont{v}}}

\newcommand{\matFont}[1]{\mathbf{#1}}

\def\mL{{\matFont{L}}}
\def\mM{{\matFont{M}}}

\def\mW{{\matFont{W}}}
\def\mX{{\matFont{X}}}
\def\mY{{\matFont{Y}}}
\def\mZ{{\matFont{Z}}}

\icmltitlerunning{An Intrinsic Vector Heat Network}

\begin{document}

\twocolumn[
\icmltitle{An Intrinsic Vector Heat Network}



\icmlsetsymbol{equal}

\begin{icmlauthorlist}
\icmlauthor{Alexander Gao}{rbx,umd}
\icmlauthor{Maurice Chu}{rbx}
\icmlauthor{Mubbasir Kapadia}{rbx}
\icmlauthor{Ming C. Lin}{umd}
\icmlauthor{Hsueh-Ti Derek Liu}{rbx}
\end{icmlauthorlist}

\icmlaffiliation{umd}{Department of Computer Science, University of Maryland, College Park, USA}
\icmlaffiliation{rbx}{Roblox}

\icmlcorrespondingauthor{Alexander Gao}{awgao@umd.edu}

\icmlkeywords{Machine Learning, ICML}

\vskip 0.3in
]



\printAffiliationsAndNotice{}  

\begin{abstract}
Vector fields are widely used to represent and model flows for many science and engineering applications.  This paper introduces a novel neural network architecture for learning tangent vector fields that are intrinsically defined on manifold surfaces embedded in 3D.  Previous approaches to learning vector fields on surfaces treat vectors as multi-dimensional scalar fields, using traditional scalar-valued architectures to process channels individually, thus fail to preserve fundamental intrinsic properties of the vector field.  The core idea of this work is to introduce a trainable vector heat diffusion module to spatially propagate vector-valued feature data across the surface, which we incorporate into our proposed architecture that consists of vector-valued neurons.  Our architecture is invariant to rigid motion of the input, isometric deformation, and choice of local tangent bases, and is robust to discretizations of the surface.  We evaluate our Vector Heat Network on triangle meshes, and empirically validate its invariant properties.  We also demonstrate the effectiveness of our method on the useful industrial application of quadrilateral mesh generation.
\end{abstract}
\section{Introduction}
Tangent vector fields on Riemannian manifolds are a fundamental ingredient in scientific computation, with applications ranging from modeling physical processes on earth \cite{sabaka2010mathematical}, to robotic navigation on complex terrains \cite{BergLM08navigation}, to mesh generation \cite{GoesDT16vectorfieldssurvey}.  The majority of works on \emph{learning} of tangent vector fields rely on neural network architectures that consist of \emph{scalar} neurons (e.g., the standard multilayer perceptron).
\begin{figure}[t]
  \centering
    \includegraphics[width=\linewidth]{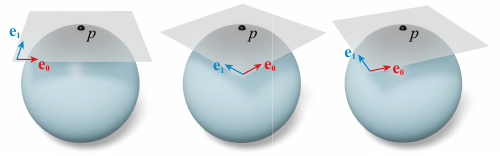}
    \caption{The tangent plane $T_p \M$ at point $p$ on a manifold $\M$ does not have a canonical choice of basis vectors $\ve_0, \ve_1$. Our proposed architecture for learning tangent vector fields is invariant to choice of tangent bases. }
    \label{fig:different_tangent_bases}
    \vspace{-2em}
\end{figure}
Despite being straightforward to implement, these scalar-valued architectures assume that each scalar channel can be processed {\em independently} of the others (e.g. as with RGB color channels in images).
However, the multiple channels of a directional vector must be considered jointly rather than independently, as they represent entangled properties (length and orientation).  As an example, a simple rotation requires adjusting all channels of the vector to maintain its length; a scalar-valued neural network does not inherently disentangle such vector properties.
Scalar-valued architectures also disregard fundamental invariances of tangent vector fields, such as the choice of local tangent bases in which the vectors are expressed (see \reffig{different_tangent_bases}).
This prohibits their generalization to unseen data that does not share the same (arbitrary) choice of bases.

In this work, we present a neural network architecture for processing tangent vector fields defined on 2-manifolds embedded in $\R^3$.
The key idea is to maintain vector-valued features throughout the architecture, and utilize a trainable vector heat diffusion process (not to be confused with Diffusion Models \cite{diffusion_models_survey}) to ensure our architecture maintains necessary invariances for tangent vector field processing.
We begin by reviewing necessary background on tangent vector field processing in \refsec{preliminaries}.  In \refsec{vector_heat_diffusion_network}, we detail our proposed architecture. In \refsec{experiments_invariances}, we show that our approach is invariant to choice of local tangent bases, rigid transformation, and isometric deformation of the input, and is robust to different {\em discretizations} of the manifold surface.
Finally, in \refsec{experiments_quad_meshing}, we highlight an application of our work in quadrilateral mesh generation for animation.

\section{Related Work}
Our work is an instance of geometric deep learning \cite{gdl_survey} focused on learning tangent vector fields defined on discrete surfaces embedded in $\R^3$. 
Such surfaces can be represented as point clouds \cite{point_cloud_survey}, implicits \cite{neural_fields_survey}, or analytical functions \cite{CohenWKW19} for simple shapes, e.g. a sphere. 
While other researchers have focused on learning vector fields defined on the entire volume, such as \cite{YangLCZ23NeuralVectorFields}, we focus on tangent vector fields defined on the \emph{triangle mesh}, a widely used surface representation for graphics, scientific computing, and engineering applications.

A majority of works on this subject focus on developing fundamental operators in neural networks (such as convolutions) to process \emph{scalar} fields defined on triangle mesh elements, such as vertices \cite{spiralnet++, meshwalker}, edges \cite{hanocka2019meshcnn, neuralsubdiv, halfedgecnn}, and faces \cite{FengFYZG19, deep_geo_texture, subdivnet}.
As a triangle mesh is merely a graph with triangular faces, graph neural networks (GNNs) \cite{WuPCLZY21GNNsurvey} have also been deployed to learning tasks on meshes, such as \cite{primal_dual_mesh_conv, PfaffFSB21}. 
Despite being effective, the fact that these operations are defined based on the \emph{connectivity} of the mesh makes them sensitive to the quality of the triangulation.

In lieu of this, spectral approaches \cite{wiersma2022deltaconv, Sharp2022diffusionnet} operate on the \emph{functional space} characterized by differential operators.
Network architectures that operate on these function spaces, such as the eigenvectors of the Laplace operator, are more robust to discretization and are able to generalize to other domains (e.g., point clouds) as long as the differential operators are defined.

The basic building block of our architecture is the \emph{vector} heat diffusion process with the \emph{connection Laplacian}.
This shares similarities with GNN architectures motivated by diffusion processes. As pointed out in \citet{sheafNN}, the graph convolution can be viewed as heat diffusion with the connection Laplacian using \emph{forward Euler} integration.
This perspective motivates the development of \emph{Sheaf Neural Networks} \cite{HansenG19, BodnarGCLB22SheafDiffusion, BattiloroWRLR23TangentBundleFilters} which learn a graph- and task-specific connection Laplacian to improve expressiveness of GNNs.
But given domain knowledge of the underlying graph (e.g., a graph that approximates a Riemannian manifold), \citet{BarberoBBBVL22sheafNN} demonstrate that a pre-determined connection Laplacian could lead to better performance and generalization.
Accordingly, since our method indeed focuses on a specific type of graph, manifold triangle meshes, we build a connection Laplacian derived from differential geometry \cite{sharp2019vector}. 
Our architecture with a deterministic connection Laplacian leads to superior generalization across triangle meshes, compared to approaches that rely on learning graph-specific Laplacians.

Despite the existence of several learning approaches for scalar data, few approaches have been proposed for learning vector fields on surfaces. 
Previous methods such as \cite{dielen2021learning} rely on scalar-valued architectures to output multiple scalar channels that are naively interpreted as vectors. Such approaches treat each channel independently and thus fail to capture key invariances (see \refsec{experiments_invariances}). This severely hinders generalization to unseen triangulations and shapes.
In contrast, our architecture maintains the vector-valued features throughout the forward pass, similar to \cite{DengLDPTG21vectorneurons}, ensuring invariance to isometry, rigid motion, and choice of tangent bases. 

\section{Preliminaries}\label{sec:preliminaries}
\begin{figure*}
\centering
    \includegraphics[width=\linewidth]{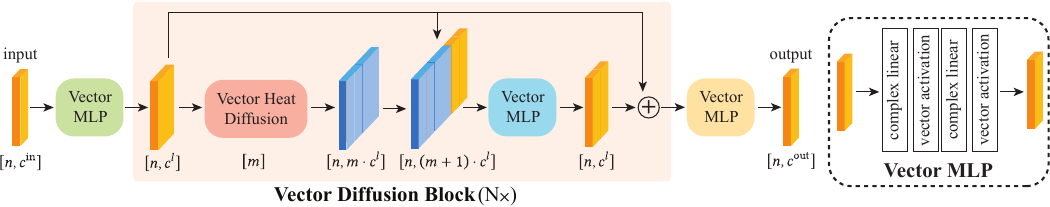}
    \caption{Our vector heat network is a neural network of complex-valued neurons \cite{complexNNsurvey} with (1) a Vector Heat Diffusion module (see \refsec{learned_diffusion}) and (2) a vector MLP module (see \refsec{complex_linear}). Starting with a Vector MLP to transform input features from $\C^{n \times c^\text{in}}$ to $\C^{n \times c^l}$, our method consists of several layers of the Vector Heat Diffusion (red) and Vector MLP (blue) with skip connections, followed by another Vector MLP to map the feature to output dimensions.} 
    \label{fig:architecture_diagram}
\end{figure*}
We draw inspiration from differential geometry to define the building blocks of our architecture, based on the notion of tangent vectors (Sec.~\ref{sec:tang_vec}), parallel transport (Sec.~\ref{sec:parallel_transport}), and the heat equation (Sec.~\ref{sec:heat}), which we describe next.

\subsection{Basis Invariant Tangent Vectors}\label{sec:tangent_vector_field}
\label{sec:tang_vec}

\begin{wrapfigure}[8]{r}{1.1in}
    \vspace{-12pt}
	\includegraphics[width=\linewidth, trim={3mm 0mm 0mm 0mm}]{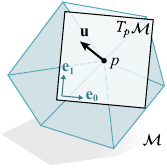}
	\label{fig:tangent_plane}
\end{wrapfigure}
Given a 2-manifold surface $\M$ embedded in $\R^3$, the tangent plane $T_p^M$ 
for a given point $p \in \M$ is a 2D space that is orthogonal to the surface normal of $\M$ at point $p$.  
A tangent plane $T_p^M$ can be defined by the spanning of an \emph{arbitrary choice} of two orthogonal basis vectors $\ve_0, \ve_1$. A 2D tangent vector $\vu \in T_p^M$ can then be expressed as a linear combination of bases vectors $\vu = u_0 \ve_0 + u_1 \ve_1$.
For computational convenience, these tangent vectors $\vu$ are often represented as a complex number $\vu = u_0 + \i u_1 \in \C$ of their coefficients $u_0, u_1$ \cite{knoppel2013globally, VaxmanCDBHBP16}. 
Note that the coefficients $u_0, u_1$ will change, by a coordinate transformation, depending on the choice of tangent bases $\ve_0, \ve_1$ in order to represent the same tangent vector. 
Since the choice of bases is arbitrary, an important invariance for tangent vector processing is to guarantee that the method is independent of the choice of bases (see \reffig{different_tangent_bases}).

\subsection{Parallel Transport of Tangent Vectors}
\label{sec:parallel_transport}
\begin{wrapfigure}[7]{r}{1.1in}
    \vspace{-12pt}
	\includegraphics[width=\linewidth, trim={3mm 0mm 0mm 0mm}]{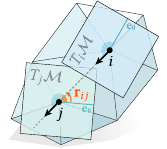}
	\label{fig:parallel_transport}
\end{wrapfigure}
An edge vector $\ve_{ij} = \vv_j - \vv_i \in \R^3$ between two adjacent vertices $i,j$ with locations $\vv_i, \vv_j \in \R^3$ can be expressed as a tangent vector in $T_i^M$ and $T_j^M$ via the \emph{logarithmic} map (sometimes alternately referred to as the \emph{exponential} map) \cite{SchmidtGW06}. 
A simple way to compute the logarithmic map is to represent the edge vector $\ve_{ij}$ in the polar coordinate $(l_{ij}, \theta_j)$ of the tangent plane $T_i^M$, where $l_{ij} = \| \ve_{ij} \|$ is the edge length and $\theta_j \in [0, 2\pi)$ denotes the angular coordinate (normalized to $2\pi$) of this edge vector from a (arbitrarily chosen) tangent basis $\ve_0$, see \cite{KnoppelCPS13} for more details. 
As the edge vector $\ve_{ij}$ exists in both tangent planes $T_i^M, T_j^M$, one can compute the angular difference between $T_i^M, T_j ^M$ and obtain the coordinate transformation $\vr_{ij} \in \C$ (a rotation) needed to make sure that $log_i(\ve_{ij}) \in T_i^M$ is mapped to $log_j(\ve_{ij}) \in T_j^M$ (see inset). 
Transporting tangent vectors from $T_i ^M$ to $T_j ^M$ following the computed rotation $\vr_{ij}$ leads to as-parallel-as-possible transport, or \emph{parallel transport}.  Note that only the angle of these vectors varies as we traverse the curved surface; their length remains unchanged.  This motivates constructing a network that can preserve these disentangled vector properties.  

\subsection{Vector Heat Diffusion}
\label{sec:heat}
\begin{figure}[!b]
  \centering
    \includegraphics[width=\linewidth]{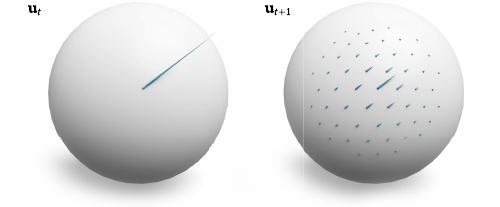}
    \caption{The vector heat diffusion process presented in \refequ{implicit_heat_diffusion} smears out a tangent vector field $\vu_t$ to its neighbors to obtain another tangent vector field $\vu_{t+1}$.  }
    \label{fig:diffuse_vector_mesh}
\end{figure}

The heat equation on a tangent vector field $u: \M \rightarrow \C$ can be written as
\begin{align}\label{equ:vector_heat_equation}
    \frac{d}{dt} u = \Delta_c u.
\end{align}
where $\Delta_c$ is the \emph{connection Laplacian}.
This vector heat equation characterizes how a ``vector'' gets diffused in space. Intuitively, diffusing a single vector heat source will smear it out to its neighborhood with smaller magnitude while maintaining its direction to be as-parallel-as-possible (see \reffig{diffuse_vector_mesh}). 
This is different from classic heat diffusion where no notion of parallel transport is captured and can cause inconsistent vector directions if one naively applies scalar heat diffusion to each vector field channel independently.

\paragraph{Discretization}
\begin{wrapfigure}[7]{r}{1.1in}
    \vspace{-12pt}
	\includegraphics[width=\linewidth, trim={3mm 0mm 0mm 0mm}]{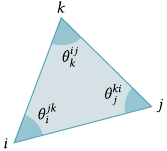}
	\label{fig:vector_cotan_weights}
\end{wrapfigure}
Several previous works \cite{KnoppelCPS15stripe, sharp2019vector, SteinWJG20} have defined the vector heat equation on a triangle mesh. 
The main ingredient is a \emph{discrete} connection Laplacian $\Lc \in \C^{n\times n}$ which is an $n$ x $n$ complex-valued matrix, where $n$ denotes the number of vertices in the mesh. 
One way to build $\Lc$ is by accumulating a 3-by-3 complex matrix
\begin{align}\label{equ:discrete_connection_laplacian}
    -\frac{1}{2} 
    \begin{bmatrix}
        c_j + c_k & - c_k \vr_{ij} & -c_j \vr_{ik} \\ 
        -c_k \vr_{ji} & c_k + c_i & -c_i \vr_{jk} \\
        - c_j \vr_{ki} & - c_i \vr_{kj} & c_i + c_j
    \end{bmatrix} 
\end{align}
associated to each triangle $ijk$ into the corresponding entries defined by the vertex index. 
We use $c_i = \cot \theta^{jk}_i, c_j = \cot \theta^{ki}_j, c_k = \cot \theta^{ij}_k$ to shorten the expression, and $\vr_{ij} \in \C$ (so as $\vr_{jk}, \vr_{ki}$) to denote a the 2D rotation (represented as a unit complex number) that \emph{parallel transports} a tangent vector from the tangent plane $T_i ^M$ at vertex $i$ to the tangent plane $T_j ^M$ at vertex $j$ (see \refsec{parallel_transport}).

As the \emph{forward Euler} method is well-known to be unstable under large time steps, we compute the numerical solution to the vector heat equation using the \emph{implicit Euler} method. Specifically, a single step of the diffusion is defined as
\begin{align}\label{equ:implicit_heat_diffusion}
    \vu_{t+1} = (\mM +  s \Lc)^{-1} \mM \vu_t
\end{align}
where $s$ to denotes the time-step size, $\mM \in \C^{n \times n}$ is a $n$ x $n$ diagonal mass matrix whose entries are complex-valued, with vertex area as the real component, and zero imaginary component. 
\refequ{implicit_heat_diffusion} diffuses a vector into smaller magnitudes and as-parallel-as-possible orientations (see \reffig{diffuse_vector_mesh}).

\subsection{Generalization to N-Rosy Fields}\label{sec:n_rosy_fields}
\begin{figure}[h]
\centering
    \includegraphics[width=\linewidth]{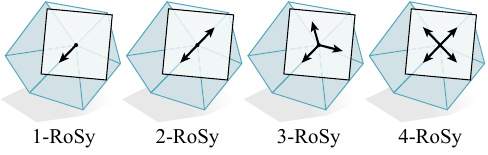}
    \caption{N-Rosy fields refer to tangent vectors that are \emph{N-way rotationally symmetric} \cite{PalaciosZ07rosy}. For instance, $N=1$ refers to the usual 2D tangent vector, $N=2$ to a straight line, and $N=4$ to a ``cross'' field.}
    \label{fig:n_rosy}
\end{figure}

In several applications, one may want to learn a \emph{N-way rotational symmetry} fields \cite{PalaciosZ07rosy} that are invariant under rotation of an integer multiple of $\nicefrac{2\pi}{N}$ (see \reffig{n_rosy}). For instance, one of our applications in quadrilateral meshing requires to output a 4-Rosy fields in order to mesh a surface with the ``cross'' pattern on most vertices. 

The generalization to N-Rosy fields is straightforward given the complex number representation (\refsec{tangent_vector_field}). Since multiplication with (unit) complex numbers represents rotations, raising a complex number to the power of $N$ factors out all the N-ways rotational symmetry \cite{GoesDT16vectorfieldssurvey}. 
Thus, to measure the difference between, e.g., 4-Rosy fields, one simply measures the difference between $\vu^4$ (see \refsec{experiments_quad_meshing}). 

\section{Vector Heat Network}\label{sec:vector_heat_diffusion_network}
Our vector-valued neural network for processing tangent vector fields utilizes vector-valued neurons (similar to \cite{DengLDPTG21vectorneurons, complexNNsurvey}) with vector operations (e.g., parallel transport and the vector heat equation).  Maintaining the vector nature of our data throughout results in an architecture that is invariant to isometries, rigid transformations, and the choice of tangent bases (see \refsec{experiments_invariances}) 

\begin{figure*}[t]
  \centering
    \includegraphics[width=\linewidth]{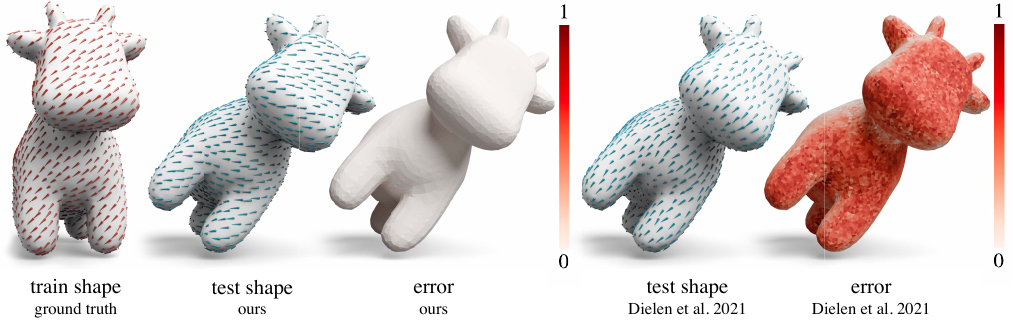}
    \caption{Our architecture is invariant to rigid transformation. A model trained on a mesh at one orientation (1st) generalizes to its rigidly transformed counterpart (2nd), outputting a tangent vector field with no error (3rd). This differs from the baseline method of \citet{dielen2021learning}, which outputs a different vector field (4th) with high error (5th).} 
    \label{fig:invariance_rigid_motions}
\end{figure*}

The input to our network is a set of tangent vectors defined on the vertices of a surface triangle mesh. 
A Vector Heat Network consists of several blocks of learned heat diffusion to harness local information, and \emph{vector MLPs} to increase the expressiveness (see \reffig{architecture_diagram}). 
The output is a set of tangent vector fields defined on vertices. These output fields could be regular tangent vectors or tangent vectors with $N$ rotational symmetry (see \refsec{n_rosy_fields}) defined on the tangent plane. 
In this section, we will illustrate individual components of our \textbf{Vector Heat Network} in more details.

\subsection{Learned Vector Heat Diffusion}\label{sec:learned_diffusion}
Starting with a set of tangent vectors $\vu \in \C^{n \times c^l}$ (represented as complex numbers, see \refsec{tangent_vector_field}), where $n$ denotes the number of vertices and $c^l$ denotes the number of tangent vector fields.
We harness information from local neighborhoods by solving the vector heat equation with the implicit Euler integration in \refequ{implicit_heat_diffusion}, inspired by \cite{Sharp2022diffusionnet}.
As the implicit Euler requires an expensive step of solving a linear system, we utilize spectral acceleration \cite{DonatiCMO22complexfmap} to speed up the process.
Specifically, given the solution of the generalized eigenvalue problem of the discrete connection Laplacian $\Lc$ (see \refequ{discrete_connection_laplacian}),
\begin{align}
    \Lc \Phi = \mM \Phi \Sigma
\end{align}
where $\Phi = \{\Phi_i\} \in \C^{n \times k}$ is the stack of $k$ eigenvectors with the lowest frequencies and $\Sigma = \textit{diag}(\lambda_i) \in \C^{k \times k}$ is a diagonal matrix with corresponding eigenvalues. We use $\mM$ to denote the mass matrix of vertex areas such that $\Phi_i^\top \mM \Phi_i = 1$. $k$ is a user-defined number to specify how many eigenvalues/eigenvectors are in use.
Then the vector diffusion process in \refequ{implicit_heat_diffusion} can be approximated with
\begin{align}\label{equ:spectral_heat_diffusion}
    \vu_{t+1} = \Phi 
    \begin{bmatrix}
        e^{- \lambda_1 s}  \\ e^{- \lambda_2 s} \\ \vdots \\  e^{- \lambda_k s} 
    \end{bmatrix}
    \odot (\Phi^\top \mM \vu_t)
\end{align}
where $\odot$ denotes element-wise multiplication.
Such a spectral acceleration replaces linear solves with matrix multiplications, thus is significantly faster for small $k$. In our implementation, we set $k = 128$. 

Inspired by \cite{Sharp2022diffusionnet}, we treat time-step size $s$ in \refequ{spectral_heat_diffusion} as trainable parameters. 
Intuitively, the network learns whether to diffuse the vectors over a small or large local neighborhood.
Specifically, each \emph{Vector Heat Diffusion} module (\reffig{architecture_diagram}) consists of $m$ trainable time steps. Each time step $s_i$ will diffuse the input feature $\mX^l \in \C^{n\times c^l}$ at layer $l$ into a set of diffused features $\mY^l_i \in \C^{n\times c^l}$ via \refequ{spectral_heat_diffusion}. Thus, a collection of $m$ time steps $[s_1, \cdots, s_m]$ will turn an input feature $\mX^l$ with size $n$ x $c^l$ into a set of diffused features $\mY^l = [\mY^l_1, \cdots, \mY^l_m]$ with size $n$ x $m_{c^l}$.

\subsection{Vector Linear Layers and Non-linearity}\label{sec:complex_linear}
After the vector heat diffusion module, we use a vector-valued MLP that consists of a per-vertex linear layer
\begin{align}
    \mZ^l = \mY^l \mW^l
\end{align}
where $\mW^l \in \R^{mc^l \times c^{l+1}}$ is a matrix of size $mc^l$-by-$c^{l+1}$ that linearly combine the complex-valued features at each vertex into $\mZ^l \in \C^{n \times c^{l+1}}$. 
Then we follow the idea presented by \cite{wiersma2022deltaconv} to apply non-linearities $\sigma$ (e.g., ReLU) on the magnitude of each complex feature as
\begin{align}
    \mX^{l+1}_{ij} = \sigma(\| \mZ^l_{ij}\| - b^l_j) \cdot \frac{\mZ^l_{ij}}{\| \mZ^l_{ij}\|}
\end{align}
where we use $\mZ^l_{ij}$ to denote the entry corresponding to the $i$th row and the $j$th column in $\mZ$, $b^l_j \in \R$ is a bias term for each channel added to the feature norm. 
In our experiments, we use the ReLU activation -- if the complex feature norm $\| \mZ^l_{ij}\|$ is smaller than the bias $b^l_j$, the complex feature is set to $\mathbf{0}$, otherwise it is unchanged in the output $\mX^{l+1}$.

In summary, our overall architecture (see \reffig{architecture_diagram}) consists of several \textbf{Vector Diffusion Blocks}. Each block contains a vector heat diffusion layer (see \refsec{learned_diffusion}) and a vector-valued MLP with two hidden layers (see \refsec{complex_linear}). 
We also have two extra vector MLPs (green, yellow) to adapt to a different number of input/output channels.
Our architecture is invariant to choice of tangent bases due to the Vector Heat Diffusion module baking in parallel transport (see \refsec{parallel_transport}).
Furthermore, the vector heat diffusion process is \emph{intrinsic}, ensuring the architecture is also invariant to rigid transformations and isometries of the underlying shape.
These desirable invariances result in a general network architecture for processing tangent vector fields on manifolds.

\section{Experiment: Invariance Properties}\label{sec:experiments_invariances}
Our architecture possesses several fundamental invariances, which we highlight and empirically validate here, distinguishing it from the scalar-valued approach of \cite{dielen2021learning}.  For comparison, we faithfully re-implement their method (\refsec{baseline_implementation}).  These invariances arise from the fact that all of our operations (gradient, heat diffusion, and the per-vertex linear layer) are \emph{intrinsic}, which implies that our architecture is invariant to how the mesh sits in the space.

\paragraph{Invariance to Rigid Motion}
If rigid motion invariant input features are used, then our method will be invariant to rigid transformations of the underlying shape (see \reffig{invariance_rigid_motions}).

\begin{figure}[h]
  \centering
    \includegraphics[width=\linewidth]{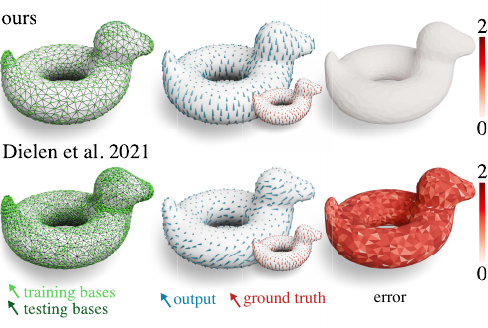}
    \caption{Our method is invariant to choice of tangent bases. Given a model trained on the default bases (light green), where the arrow indicates direction of \textbf{e0}, our method produces the same result (blue) as GT (red) even when the model is evaluated under different choice of bases (dark green), in contrast to \cite{dielen2021learning} that outputs a different result with high error (right).} 
    \label{fig:invariance_local_basis}
\end{figure}

\paragraph{Invariance to Tangent Bases} We leverage the characteristic that parallel transport has already factored out the influence of choice of tangent bases, which is {\em baked in} to our connection Laplacian. This property makes our architecture  invariant to the choice of tangent bases (\reffig{invariance_local_basis}).

\begin{figure*}
  \centering
    \includegraphics[width=\linewidth]{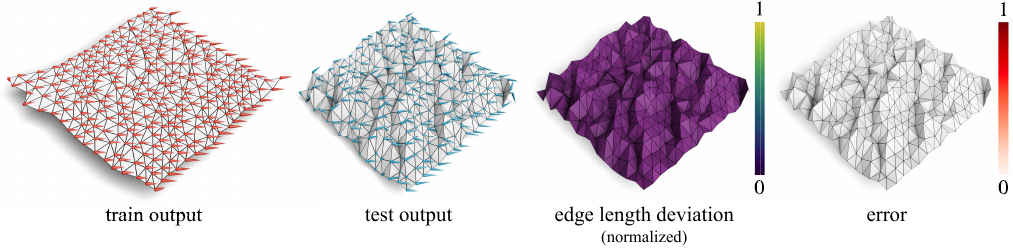}
    \caption{Our method is invariant to isometric deformation. We train on a flat paper (first) and evaluate on its crumpled counterpart (second). Since these two meshes are nearly isometric (third), our method produces consistent results (fourth).} 
    \label{fig:isometry}
\end{figure*}

\paragraph{Invariance to Isometry}
Our method is invariant to isometric deformation (isometries) of the input (see \reffig{isometry}), due to the intrinsic construction of our method.

\paragraph{Robustness to Discretizations} Our main ingredient for ``message passing'' relies on the vector heat diffusion with the connection Laplacian. In contrast to the method by \citet{BodnarGCLB22SheafDiffusion} that learns the connection Laplacian for a graph, the connection Laplacian in our set-up is \emph{deterministic} by the underlying shape, with \emph{connections} determined by the parallel transport (\refsec{parallel_transport}). This characteristic ensures that our learned parameters are generalizable to meshes with different connectivities. This property is crucial to tangent vector field processing because the choice of tangent bases is arbitrary (see \reffig{different_tangent_bases}): one can find an infinite number of \emph{valid} tangent bases $\ve_0, \ve_1$ that are orthogonal to the normal vector. On a single vertex, there is already an infinite number of choices and the total combinations of basis choices also grows exponentially with the number of vertices in the mesh. 
This implies that baking in the property of basis invarance is important because solving it with data augmentation is intractable due to the infinite number of basis combinations.

\begin{figure}[h]
  \centering
    \includegraphics[width=\linewidth]{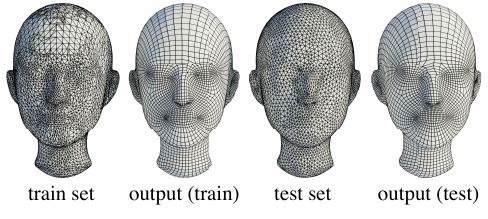}
    \caption{This work exploits the benefit of the spectral method (see \refequ{spectral_heat_diffusion}) and gains robustness to different discretizations. Trained on one triangulation (1st), our model generalizes to a different one (3rd), producing consistent output (quad meshing results from \refsec{experiments_quad_meshing}) shown here.)}
    \label{fig:robustness_triangulations}
    \vspace{-1em}
\end{figure}

\paragraph{Baseline Implementation Details}\label{sec:baseline_implementation}
Our re-implementation of the architecture by \citet{dielen2021learning} is as consistent as possible with theirs: for the local feature network (\cite{spiralnet++}), we use a spiral sequence length of $k = 20$ vertices, with $4$ spiral convolution layers of intermediate size $[16, 256, 512, 1024]$, where the first 3 layers use vertex-centric spiral indices, and the last layer uses face-centric spiral indices; the global feature (\cite{qi2017pointnet}) consists of $1024$ channels; and each input vertex is represented as its 3D position and normal direction.

\section{Experiment: Quadrilateral Remeshing}\label{sec:experiments_quad_meshing}
In this section, we evaluate our method on triangle meshes, though the same principles apply to other domains where the connection Laplacian is available (e.g. point clouds).

\subsection{Experiment Setup}
\begin{figure}[h]
  \centering
    \includegraphics[width=\linewidth]{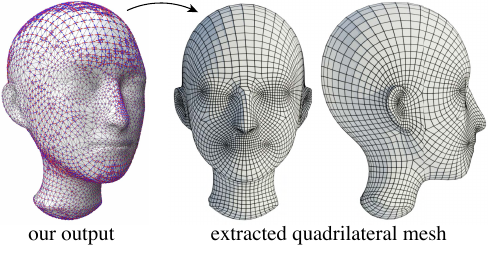}
    \caption{After obtaining the output cross field by our network, we use off-the-shelf quad mesh extraction method by \citet{bommes2009mixed} to obtain a quadrilateral mesh with edges aligned with the predicted cross field.} 
    \label{fig:quad_extraction}
\end{figure}

To demonstrate our method's effectiveness, we evaluate it on the task of quadrilateral remeshing.
Given a triangle mesh $\M$, we use the per-channel gradient of the first $c^\text{in}=15$ channels of the Heat Kernel Signature (HKS) \cite{sun2009concise} as input features to our network, giving $c^\text{in}$ vector-valued features per vertex. 
The output is a 4-Rosy cross field defined on each vertex (see \refsec{n_rosy_fields} and \reffig{quad_extraction}). 
We then interpolate the per-vertex cross field onto faces (see \refsec{verts2faces}), followed by off-the-shelf algorithms by \citet{bommes2009mixed} and \cite{ebke2013qex} to turn the per-face cross field into a quadrilateral mesh (see \reffig{quad_extraction}).

\paragraph{Dataset} 
We train our network on a dataset generated from the workflow described in \cite{dielen2021learning}, with two modifications: (1) Instead of the DFAUST dataset used by \cite{dielen2021learning}, we assemble a custom library of artist-created template avatar heads, around which we wrap the SMPL \cite{loper2023smpl} head topology; (2) for each of the template meshes, we create $100-1,000$ augmentations/variations, using a custom tool for deforming faces, based upon normal-driven ARAP deformation \cite{LiuJ21}. This expands the training data distribution away from the parametric SMPL model. The training dataset consists of $1100$ triangle meshes with associated ground truth vector fields. The test dataset consists of $115$ samples.

\paragraph{Loss Function}\label{para:loss_function}
Since our output is a 4-Rosy field, we define the loss function as the mean squared error (MSE) on the output tangent vector field $\vu^4$ raised to the power of 4 (see \refsec{n_rosy_fields}). In addition to measuring errors on the directions, we also want to measure errors on the magnitude of the cross field (smaller crosses lead to smaller polygon). Combining the two leads to our following loss function
\begin{equation}
\begin{aligned}\label{equ:size_direction_loss}
    &\mathcal{L}(\vu, \hat{\vu}) = \\
    &\sum_{i = 1}^N \frac{\textbf{M}_{ii}}{A} \Bigg[\underbrace{\bigg|\frac{\|u_i\| - \|\hat{u_i}\|}{\|u_i\|}\bigg|}_\text{magnitude} +
    \underbrace{\bigg(1 - \frac{u_i^4}{\|u_i^4\|}\cdot\frac{\hat{u_i}^4}{\|\hat{u_i}^4\|}\bigg)}_\text{direction}\Bigg]
\end{aligned}
\end{equation}

where $\mM_{ii}$ denotes the vertex area at vertex $i$, $A = \Tr(\mM)$ is the total area of the mesh, $u_i, \hat{u_i} \in \C$ denote the output and the ground truth tangent vector fields on the vertex $i$, respectively.
The first ``magnitude'' term simply measure the relative difference in magnitude between the output and the ground truth fields. The second ``direction'' term measures their angular difference with the cosine similarity.

\paragraph{Transporting Vectors from Vertices to Faces}\label{sec:vertex_to_face_transport}\label{sec:verts2faces}
Our model predicts vectors at mesh vertices, but in order to directly leverage existing methods for field-guided quadrilateral remeshing \cite{bommes2009mixed}, they should be expressed on the face tangent planes.  Naively averaging the three vector predictions from a given face's three incident vertices will not produce a correct result, as the vectors are expressed with respect to their individual vector tangent planes, so they cannot be averaged directly.  We must therefore account for the parallel transport from each of the vertex tangent planes to the face tangent plane.  Once all three vectors are in a shared frame, then we may simply average their values. 

To transport a vector from its tangent plane $T_i \M$ at a vertex $v_i$ to the tangent plane $T_{ijk} \M$ at face $f_{ijk}$, we choose an edge that is incident to both elements (e.g., edge $e_{ij}$ or $e_{ik}$).  The chosen edge can be expressed in terms of its angular rotation from the local basis in both the vertex $T_i \M$ and face tangent planes $T_{ijk} \M$.  Thus, we leverage this angular difference to transport a vector from vertex to face tangent plane.  This transport can be constructed as an operator in a pre-processing step, as it only depends on the mesh, i.e. once we have computed the angular differences, they may be used to transport any vectors from vertices to faces.  Let this matrix be $\textbf{T}_{angle} \in \R^{F \times 3}$, where each row corresponds to a face, and each of the three values corresponds to the transition angle (in radians) needed to transport a vector from the incident vertex at that index, to the face.  Then we can assemble the complex matrix \emph{operator}:
\begin{equation}
    \textbf{T} = e^{\i\textbf{T}_{angle}}
\end{equation}
To compute the transported, averaged result per face,
\begin{equation}
    \hat{\textbf{u}}_{f} = \frac{1}{3}\sum\textbf{T} \odot \hat{\textbf{u}}_{v}[\textbf{F}]
\end{equation}
where the $[\cdot]$ represents an indexed selection from the per-vertex predictions based on vertex indices from the face matrix $\textbf{F}$, and the sum $\sum(\cdot)$ denotes a per-face addition along the last dimension of the product $\textbf{T} \odot \hat{\textbf{u}}_{v}[\textbf{F}]$.

\begin{figure}[t]
  \centering
    \includegraphics[width=\linewidth]{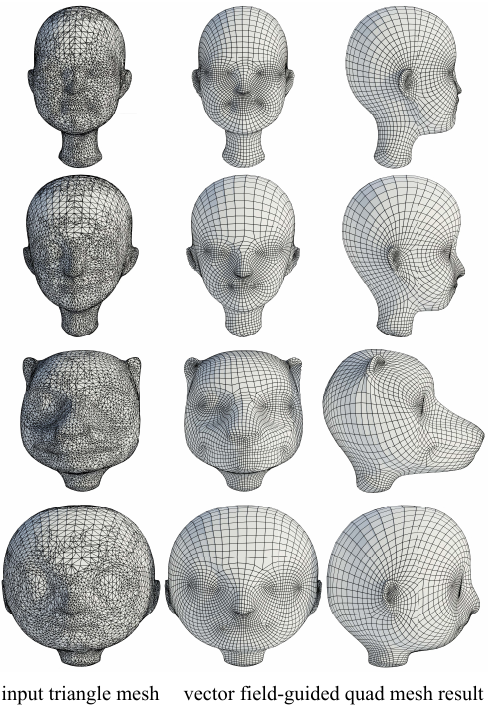}
    \caption{Vector-field-guided quadrilateral meshing results of various character heads from the test dataset.} 
    \label{fig:quad_mesh_results}
\end{figure}

\begin{figure*}[t]
  \centering
    \includegraphics[width=\linewidth]{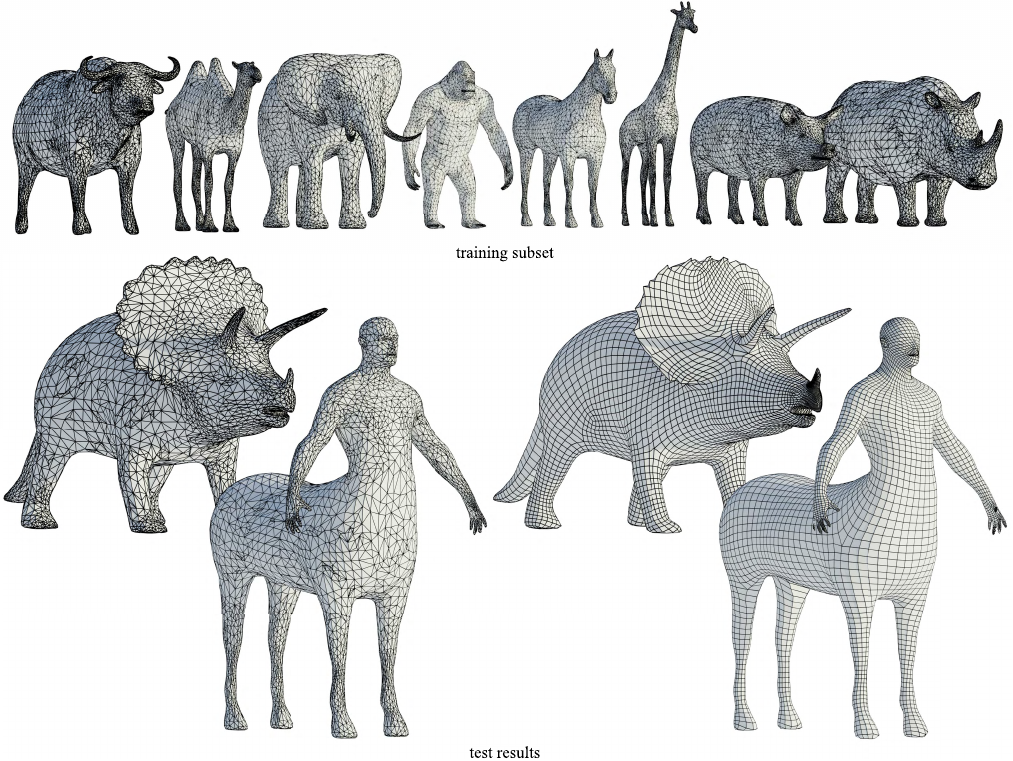}
    \caption{Our learned vector-field-guided quadrilateral remeshing can generalize to other datasets and classes of objects.} 
    \label{fig:generalization_results}
\end{figure*}

\paragraph{Implementation Details}\label{sec:implementation_details}
For our experiments, we use $N = 6$ vector diffusion blocks with a hidden dimension of $c^l = 256$ channels (see \reffig{architecture_diagram}).  We train for $3,000$ epochs, with initial learning rate of $1e-4$, decayed by a factor of $0.85$ every $150$ epochs.  In the Vector MLP layer, we use Dropout \cite{srivastava2014dropout} set to $0.5$, and L2 regularization (weight decay) with a value of $1e-3$, which mitigates overfitting.  We train on a single NVIDIA Tesla T4 GPU, for about $20$ hours.  For computing a parametrization from the predicted vector field, we rely on \cite{bommes2009mixed}, rather than \cite{campen2015quantized} used by \cite{dielen2021learning}, due to the latter not being open sourced.

\subsection{Results}
\paragraph{Character Heads} In \reffig{quad_mesh_results}, we display several results of our learned vector field-guided quadrilateral meshing on the character head dataset.  Our model generalizes across different types of character heads.

\paragraph{Animals} \reffig{generalization_results} shows results on a different dataset that we constructed from 40 ground truth quadrilateral meshes of animals.  The trained model generalizes to \emph{centaur} with its human-like upper body and horse-like lower body, even though such a combination is not contained in the training dataset.  The \emph{triceratops} result generally looks suitable for downstream usage, though its frill (collar) shows some irregularity, likely due to lack of sufficient training data.

\begin{figure}[!b]
  \centering
    \includegraphics[width=\linewidth]{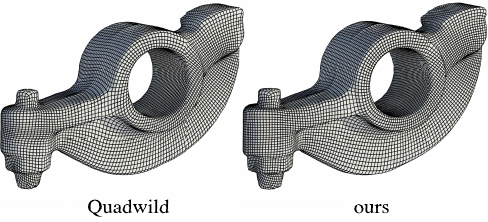}
    \caption{We supervise our method with an optimization-based quad meshing algorithm -- Quadwild by [Pietroni et al. 2021], faithfully reproducing their result.}
    \label{fig:quadwild_sm}
\end{figure}

\begin{figure}[t]
  \centering
    \includegraphics[width=\linewidth]{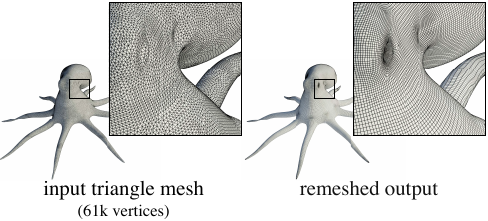}
    \caption{Our method is scalable to high resolution input meshes. We demonstrate a quad meshing result on this octopus mesh with over 60k vertices.}
    \label{fig:resolution_sm}
\end{figure}

\paragraph{Additional experimental results} Our method can faithfully reproduce the results from existing quad meshing algorithms (\reffig{quadwild_sm}), such as Quadwild \cite{pietroni2021reliable}.  We also note that our method imposes no restriction on the genus of the shape on which vector fields may be learned, as demonstrated by successful generalization to genus-one shapes in \reffig{invariance_local_basis} and \reffig{quadwild_sm}.  We also show that our method easily scales to high resolution meshes (\reffig{resolution_sm}).  Please see \refsec{additional_experiments} for enlarged detail views.

\paragraph{Ablation Studies}\label{subsec:ablation_input_features} We compare the test set performance of various types of input features, including \emph{principal curvature directions}, and the gradients of: \emph{Heat Kernel Signature} \cite{sun2009concise}, \emph{Gaussian curvature}, and \emph{mean curvature}.  For multi-channeled feature types, we normalize each of its channels individually, such that its mean vector length across all vertices is unit.  For all feature types \emph{except PCD}, we also rotate each channel by $\frac{\pi}{2}$ radians, and concatenate these rotated vector features along the channel dimension.  In principle, this means that each input feature channel and its rotated counterpart span the local tangent space, allowing the network to better exploit all degrees of freedom.

\begin{table}[h]
  \begin{tabular}{ccl}
    \toprule
    Input Feature Type & Direction Loss & Magnitude Loss  \\
    \midrule
    $\nabla$HKS & $0.106 \pm 0.278$ & $\mathbf{0.077 \pm 0.148}$ \\
    $\nabla$GC & $0.139 \pm 0.312$ & $0.096 \pm 0.261$  \\
    $\nabla$MC & $\mathbf{0.105 \pm 0.276}$ & $0.077 \pm 0.514$  \\
    PCD & $0.128 \pm 0.313$ & $0.090 \pm 0.288$  \\
  \bottomrule
\end{tabular}
\caption{Input features are evaluated by comparing their mean test loss and associated variance of their magnitude and direction components.  $\nabla$\emph{HKS} denotes channel-wise gradient of the scalar-valued \emph{Heat Kernel Signature}, $\nabla$\emph{GC} denotes gradient of \emph{Gaussian Curvature}, $\nabla$\emph{MC} denotes gradient of \emph{Mean Curvature}, and \emph{PCD} denotes scaled principal curvature directions.  We find that $\nabla$\emph{HKS} leads to best overall performance.  While $\nabla$\emph{MC} performs best on the directional loss component only, it displays high variance in the magnitude loss component.}
\label{tab:freq}
\end{table}

\section{Conclusion}
We present a neural network architecture based on vector heat diffusion to process tangent vector fields defined on manifold surfaces. Unlike existing works, our method is invariant to rigid transformations and the choice of tangent plane bases, and is robust to different triangulations. These properties jointly make this network a generalizable architecture for learning tangent vector fields across surfaces.

\begin{figure}[h]
  \centering
    \includegraphics[width=\linewidth]{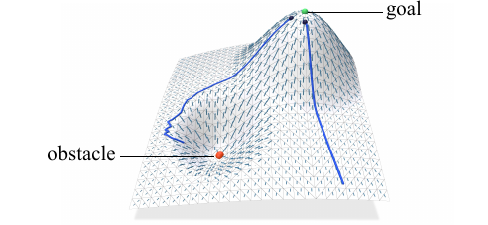}
    \caption{Tangent vector fields are commonly used to guide a robot's path planning [Patil et al. 2010]. We train our method to generate such a navigation vector field on the terrain mesh to assist motion planing as shown.}
    \label{fig:robot_navigation}
\end{figure}

\paragraph{Additional Applications}  We show an application of our method to robot path planning on curved terrain (\reffig{robot_navigation}).

\noindent
{\bf Future Directions} As the vector heat equation can be defined in different domains, such as point clouds and graphs, generalizing our architecture to different domains could enable a even wider range of applications.  Exploring novel architectures from \emph{complex neural networks} \cite{complexNNsurvey} or even generalizing to quaternions, such as \cite{ZhuXXC18}, could lead to novel variants of our architecture.
For the quadrilateral meshing application, developing a larger, more diverse dataset could be an important step towards a generic learning-based remesher.

\section*{Acknowledgements} 
Thank you to Jing Liang for help executing the robot motion planning application in ROS, Jihyun Yoon for help assembling the 3D heads dataset, and Kiran Bhat for helpful discussion.  This research is supported in part by Barry Mersky Endowed Professorship, Capital One Endowed Professorship, and Maryland E-Nnovate Initiative Fund. 

\section*{Impact Statement} The goal of this work is to advance the field of Machine Learning. There are many potential societal consequences of our work, none which we feel must be specifically highlighted here.

\newpage

\bibliography{sections/references}
\bibliographystyle{icml2024}

\newpage
\appendix
\onecolumn
\section{Additional experimental results (enlarged detail view)}\label{sec:additional_experiments}
\begin{figure}[h]
  \centering
    \includegraphics[width=\linewidth]{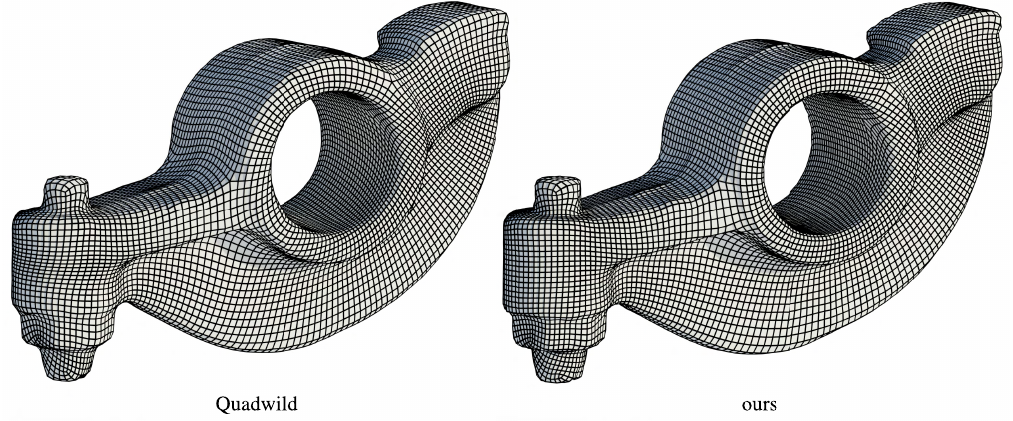}
    \caption{We supervise our method with an optimization-based quad meshing algorithm -- Quadwild by [Pietroni et al. 2021]. Our method is able to reproduce the result from Quadwild.}
    \label{fig:quadwild}
\end{figure}

\begin{figure}[h]
  \centering
    \includegraphics[width=\linewidth]{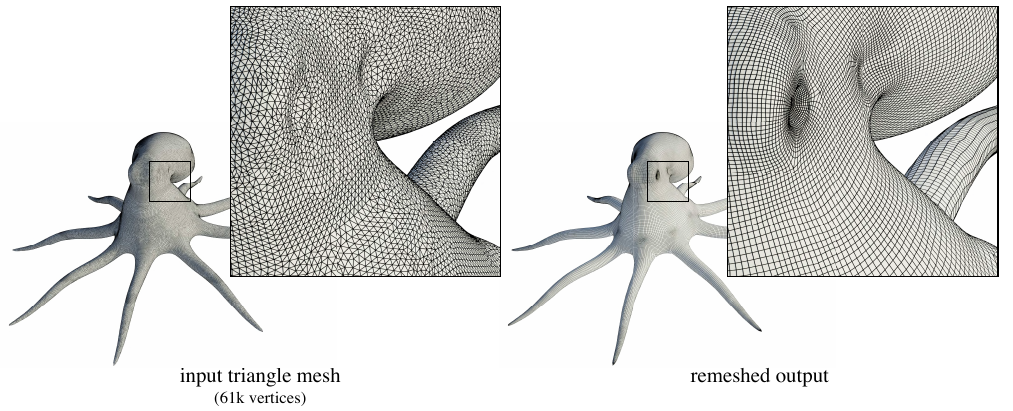}
    \caption{Our method is scalable to high resolution input meshes. We demonstrate a quad meshing result on this octopus mesh with over 60k vertices.}
    \label{fig:resolution}
\end{figure}

Our method can faithfully reproduce the results from existing quad meshing algorithms (\reffig{quadwild}), such as Quadwild \cite{pietroni2021reliable}.  We also show that our method easily scales to high resolution meshes (\reffig{resolution}).

\end{document}